%% file: root.tex
\documentclass[letterpaper, 10 pt, conference]{ieeeconf}  

\IEEEoverridecommandlockouts                              

\usepackage{url}
\usepackage{cite}
\usepackage{xcolor}
\usepackage{amsfonts}
\usepackage{graphicx}
\usepackage{amsmath}
\usepackage{booktabs}
\usepackage{makecell}
\usepackage{float}
\usepackage{tabularx}
\usepackage[font=small]{caption}
\usepackage{booktabs}
\usepackage{makecell}
\usepackage{adjustbox}
\title{\LARGE \bf CEER2: Directional and Tunable End-Effector and Root  \\ Compliance for Humanoid Loco-Manipulation}

\author{
Xinyuan Luo, Chunyuan Yang, Boyuan Chen, Xianyi Cheng%
\thanks{This work is supported by DARPA TIAMAT program under award HR00112490419 and ARO under award W911NF2410405.}%
\thanks{All authors are with the Department of Mechanical Engineering and Materials Science, Duke University, Durham, NC 27708, USA.}%
}

\IEEEaftertitletext{%
\begin{minipage}{\textwidth}
    \centering
    \includegraphics[width=0.95\textwidth]{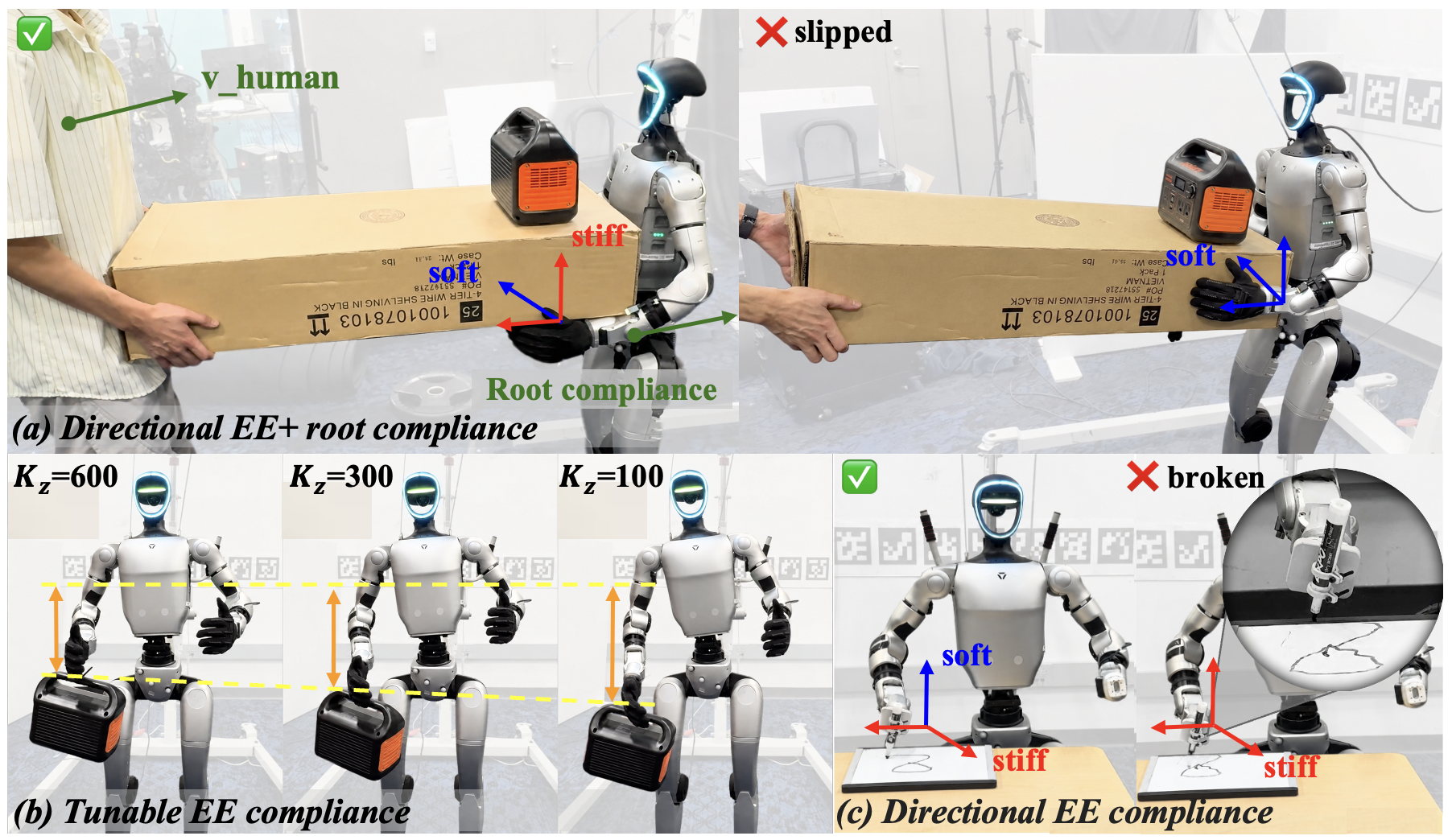}
    \captionof{figure}{Our framework enables directional and tunable EE compliance together with three root modes: resist, damping $B{=}60$, and $B{=}200$ damping. (a) Directional EE and root compliance support collaborative box carrying, while an isotropically soft setting causes the payload to slip. (b) Online changes in $K_z$ produce different vertical EE displacements while dragging a $3\,\mathrm{kg}$ payload. (c) Directional EE compliance enables accurate yet compliant writing, whereas excessive stiffness can damage the contact tool.}
    \label{fig:teaser}
    \vspace{0.5em}
\end{minipage}%
}

\begin{document}
\maketitle
\thispagestyle{empty}
\pagestyle{empty}
\bstctlcite{IEEEexample:BSTcontrol}

\begin{abstract}

Humanoids are increasingly capable of tracking complex whole-body motions, but physical interaction introduces a different challenge. When a robot makes contact with a person or the environment, it needs to respond to external forces while preserving the motion needed for the task. This response can vary across directions in the end-effectors and on the body. For example, an end effector may need to accommodate contact force in one direction while maintaining motion accuracy in another, while the robot body may resist an external force or move with it. We present a compliance framework for humanoid loco-manipulation that combines directional and tunable end-effector (EE) compliance with selectable root compliance for external force rejection or force following. A hierarchical reinforcement learning controller modulates a fixed whole-body tracking policy through high-level EE and root commands, while interaction forces are estimated from proprioceptive history. Our simulation and real-world experiments on a humanoid demonstrate directional stiffness control, online stiffness adjustment, distinct root compliance, compliant manipulation, and collaborative carrying. Our project page is at: \url{https://ee-root-compliance.github.io/}

\end{abstract}

\input{introduction}
\input{related_work}
\input{method}
\input{experiments}

\section{Conclusion}
We presented a structured formulation of humanoid compliance that combines
directional and tunable EE compliance with root damping and resistance. A
two-level hierarchical RL controller with an analytical compliance control MoE
realizes these behaviors on top of a shared stiff low-level whole-body tracking
policy,
supporting online adjustment of EE stiffness and root mode switching. Simulation
evaluations and real-world loco-manipulation
experiments show that different compliance configurations enable safer contact,
more accurate manipulation, payload adaptation, and collaborative physical
interaction.

The current implementation has several limitations. First, the achievable compliance behavior is limited by the capability of the stiff low-level whole-body tracking policy. Since the high-level controller only modulates its EE--root commands, weaknesses of the underlying tracker directly carry over to the compliance controller. In particular, insufficient $x$-direction stiffness limits the achievable EE stiffness range, while limited locomotion capability constrains root-level interaction performance. Second, although the framework supports three root compliance modes, physical interaction is not yet sufficiently smooth. The robot often takes heavy steps and lacks deliberate foot-placement adaptation under interaction. Future work will improve root compliance by explicitly incorporating interaction-dependent foot-placement targets and gait regularization that penalizes large step impacts and abrupt contact transitions.







\bibliographystyle{IEEEtran}
\bibliography{references}

\end{document}

%% file: introduction.tex
\section{INTRODUCTION}

Recent advances in humanoid whole-body control have established increasingly
capable policies that can serve as general behavioral foundations. Feature-based dense
tracking methods \cite{luo2026sonic,qi2026humanoidgptscalingdatastructure} support scalable motion tracking and provide interfaces for high-level planning, whereas generative-adversarial based models \cite{li2025bfmzeropromptablebehavioralfoundation} improve recovery and transitions across diverse motions. These capabilities allow humanoids to execute complex whole-body motions with increasing accuracy. However, during physical interaction, continuing to track the reference motion without responding to external forces can generate excessive interaction forces or destabilize the robot. Accurate motion tracking alone does not specify how a humanoid should respond when physically interacting with humans or the environment.

Recent compliant and force-aware humanoid controllers
\cite{softmimic,gentlehumanoid,luo2026ceer,chip,
jang2026wtumitactilebasedwholebodymanipulation,
chen2026sixthsensetaskagnosticproprioceptiononlywholebody,
shi2026minimalistcompliancecontrol}
improve physical interaction beyond motion tracking, but they do not jointly
specify directional, tunable EE compliance and root response. Simply making an
EE isotropically soft can reduce contact force, but it sacrifices accuracy in
task-relevant directions. Writing, for example, requires low stiffness normal
to the surface and high in-plane stiffness, while different contacts may also
require different stiffness magnitudes. In addition, different loco-manipulation tasks have different specifications on whether the body should resist, yield, or follow the external force. More generally, how to achieve whole-body compliance for a floating-base humanoid remains an open problem.

In this work, we take a step toward structured humanoid compliance by organizing it around two primary functional components: \textbf{manipulation} and \textbf{locomotion}. For manipulation, we introduce directional EE compliance that is continuously tunable over $100$--$600\,\mathrm{N/m}$, allowing different Cartesian stiffness values to be commanded along different axes. For locomotion, we represent compliance at the root through one resistance mode and two distinct damping modes. Together, they provide a structured interface in which EE stiffness and root response can be adjusted independently and applied concurrently during physical interaction.

Realizing these behaviors with a single policy is challenging because training
across external forces and axis-wise stiffness commands creates substantial
multi-task interference, while deployment without wrist force/torque sensors
requires force estimation. We therefore use a two-stage hierarchical RL
pipeline: a stiff low-level whole-body tracking policy is trained first and
then held fixed while high-level residual policies and a force estimator are
jointly trained through teacher--student learning with privileged force
observations. At deployment, an analytical compliance-space mixture-of-experts
(MoE) composes stiffness-specific EE experts, while a root-mode command selects
one of three root policies. This design separates compliance learning from
motion tracking and reduces interference among tasks with different stiffness requirements. Simulation
evaluations show that the hierarchy and analytical MoE improve full-range
directional EE compliance and realize distinct root responses, while real-world
tasks demonstrate online stiffness adjustment and concurrent EE--root
compliance. In summary, our contributions are:

\begin{itemize}
    \item We introduce a structured formulation of humanoid compliance spanning both manipulation and locomotion: directional and continuously tunable EE compliance for manipulation, and one resistance mode and two damping modes at the root for locomotion.

    \item We develop a two-level hierarchical RL controller in which high-level command residuals realize softer EE responses and root compliance on top of a fixed, stiff low-level whole-body tracking policy. An analytical compliance-space MoE composes 10 EE experts to provide independently adjustable stiffness along each Cartesian axis over $100$--$600\,\mathrm{N/m}$.

    \item We validate the proposed compliance definition through real-world manipulation and loco-manipulation tasks, demonstrating the utility of directional and tunable EE compliance together with root compliance in physical interaction.
\end{itemize}

%% file: related_work.tex
\section{Related Work}

\begin{table}[t]
\vspace{2mm}
\centering
\caption{Comparison of humanoid compliance methods.}
\label{tab:compliance_comparison}

\resizebox{\columnwidth}{!}{
\Large
\begin{tabular}{lcccccc}
\toprule
\textbf{Method} &
\textbf{Direction} &
\textbf{Angular} &
\textbf{Tunable} &
\textbf{Gen. Track.} &
\textbf{Scope} &
\textbf{Loco.} \\
\midrule

Gentle \cite{gentlehumanoid}
& $\times$
& $\times$
& 100--300$^\dagger$
& $\checkmark$
& Upper body
& $\times$ \\

CHIP \cite{chip}
& $\times$
& $\times$
& $1/k:~0$--$0.05^\ddagger$
& $\checkmark$
& EE
& $\times$ \\

SoftMimic \cite{softmimic}
& $\times$
& $\times$
& 40--1000
& $\times$
& EE
& $\times$ \\

CEER \cite{luo2026ceer}
& $\times$
& $\times$
& $\times$
& $\checkmark$
& EE
& $\times$ \\

LAC \cite{liu2026laclinearangularcompliance}
& $\times$
& $\checkmark$
& 10--500
& $\checkmark$
& Arms + torso
& $\times$ \\

\textbf{Ours}
& $\checkmark$
& $\times$
& \textbf{100--600}
& $\checkmark$
& \textbf{EE + root}
& $\checkmark$ \\

\bottomrule
\end{tabular}
}

\vspace{1mm}
\begin{minipage}{\columnwidth}
\footnotesize
All unmarked tunable stiffness ranges are in N/m.
$^\dagger$Gentle controls a force limit of 5--15\,N; the reported
100--300\,N/m corresponds to the reference-dynamics stiffness derived
from $K_p=\tau_{\mathrm{safe}}/0.05$, rather than measured Cartesian stiffness.
$^\ddagger$CHIP controls the compliance coefficient $1/k$ in the range
0--0.05\,m/N, rather than directly commanding Cartesian stiffness;
$1/k=0$ corresponds to the stiff limit.
\end{minipage}

\vspace{-2mm}
\end{table}

\subsection{Compliance Control for Humanoid Robots}

Recent works have introduced compliant behaviors into humanoid control from
different perspectives. GentleHumanoid~\cite{gentlehumanoid} focuses on
force-limited compliant tracking, CHIP~\cite{chip} introduces tunable
end-effector compliance, and SoftMimic~\cite{softmimic} uses IK-based compliant
motion augmentation to generate feasible whole-body reference responses,
enabling balanced and posture-preserving compliance.
CEER~\cite{luo2026ceer} provides a reusable end-effector compliance interface
built on a general whole-body tracking policy. More recently,
LAC~\cite{liu2026laclinearangularcompliance} extends compliance to multiple
upper-body links and introduces explicit angular compliance, while
Minimalist Compliance Control~\cite{shi2026minimalistcompliancecontrol}
combines estimated interaction forces with analytical task-space admittance
control.
Table~\ref{tab:compliance_comparison} summarizes these methods along several
dimensions. ``Direction'' denotes independently controllable directional
Cartesian stiffness, ``Gen. Track.'' denotes general motion-tracking
capability, ``Scope'' indicates where compliance is explicitly defined, and
``Loco.'' denotes an explicit root compliance formulation for locomotion. Our method is the only one to explicitly support directional compliance. It also provides a wide tunable stiffness range while retaining general tracking capability. More importantly, our formulation covers both end-effector manipulation and root-level locomotion responses, providing a broader and more functionally structured formulation of humanoid compliance.

\subsection{Force- and Tactile-Aware Humanoid Interaction}

Beyond explicit compliance control, recent works incorporate interaction forces
or tactile feedback into policy learning. Force-adaptive methods use
physical interaction information to improve whole-body robustness and
coordination under external forces
\cite{falcon, pudasaini2026fameforceadaptiverlexpanding,
li2026thorhumanlevelwholebodyreactions,
dong2026hafoforceadaptivecontrolframework, cola}.
For manipulation, tactile sensing and force-aware control have
also been integrated into learned humanoid policies
\cite{jang2026wtumitactilebasedwholebodymanipulation, tact,
wei2025hmclearningheterogeneousmetacontrol}.
Other works focus on estimating external contacts or wrenches from
proprioception and robot dynamics, providing interaction information that can
subsequently support force-aware control
\cite{chen2026sixthsensetaskagnosticproprioceptiononlywholebody}.
%
These approaches make force and contact information available for interaction control; however, sensing or estimating interaction does not itself specify
distinct compliance behaviors.

\subsection{Contact-Aware Humanoid Loco-Manipulation}

Complementary to methods that reason about interaction forces, contact-aware
loco-manipulation methods explicitly represent contact relationships between
the robot, objects, and the environment.
One line of work focuses on preserving contact structure during motion retargeting or imitation.
OmniRetarget~\cite{omniretarget}, HDMI~\cite{hdmi}, and InterMimic~\cite{intermimic} transfer human-object interaction patterns to humanoids while maintaining physically meaningful contact relationships.
Another line introduces contact information directly into the policy interface.
SceneBot~\cite{scenebot}, OmniContact~\cite{omnicontact}, and ContactMimic~\cite{contactmimic} condition whole-body control on explicit contact labels, schedules, or body-part contact commands, enabling runtime control over where and when contacts occur.
Geometry-aware approaches such as LessMimic~\cite{lessmimic} further encode interaction through distance fields and local geometric features rather than predefined contact labels.

Together, these lines of work advance humanoid compliance control, contact force estimation, and motion generation with contact constraints. Our work complements these methods by specifying \emph{how} the robot responds to contact through directional and tunable EE compliance and selectable root compliance.

%% file: method.tex
\section{Method}
\label{sec:method}

We use a two-level hierarchy with a low-level whole-body tracking policy and high-level compliance policies to reduce multi-task interference across compliance targets during learning.
The key idea is to use a sufficiently stiff low-level whole-body
tracking policy as the fixed backbone, allowing high-level compliance policies
to realize a range of softer behaviors through command residuals. We train
separate high-level compliance policies for different compliance settings. We
further find that the EE compliance experts can be analytically composed in
compliance space to provide continuous, independently adjustable stiffness
along each Cartesian axis, while the root compliance policies are selected
through mode switching.

\begin{figure*}[t]
\vspace{2mm}
    \centering
    \includegraphics[width=0.9\textwidth]{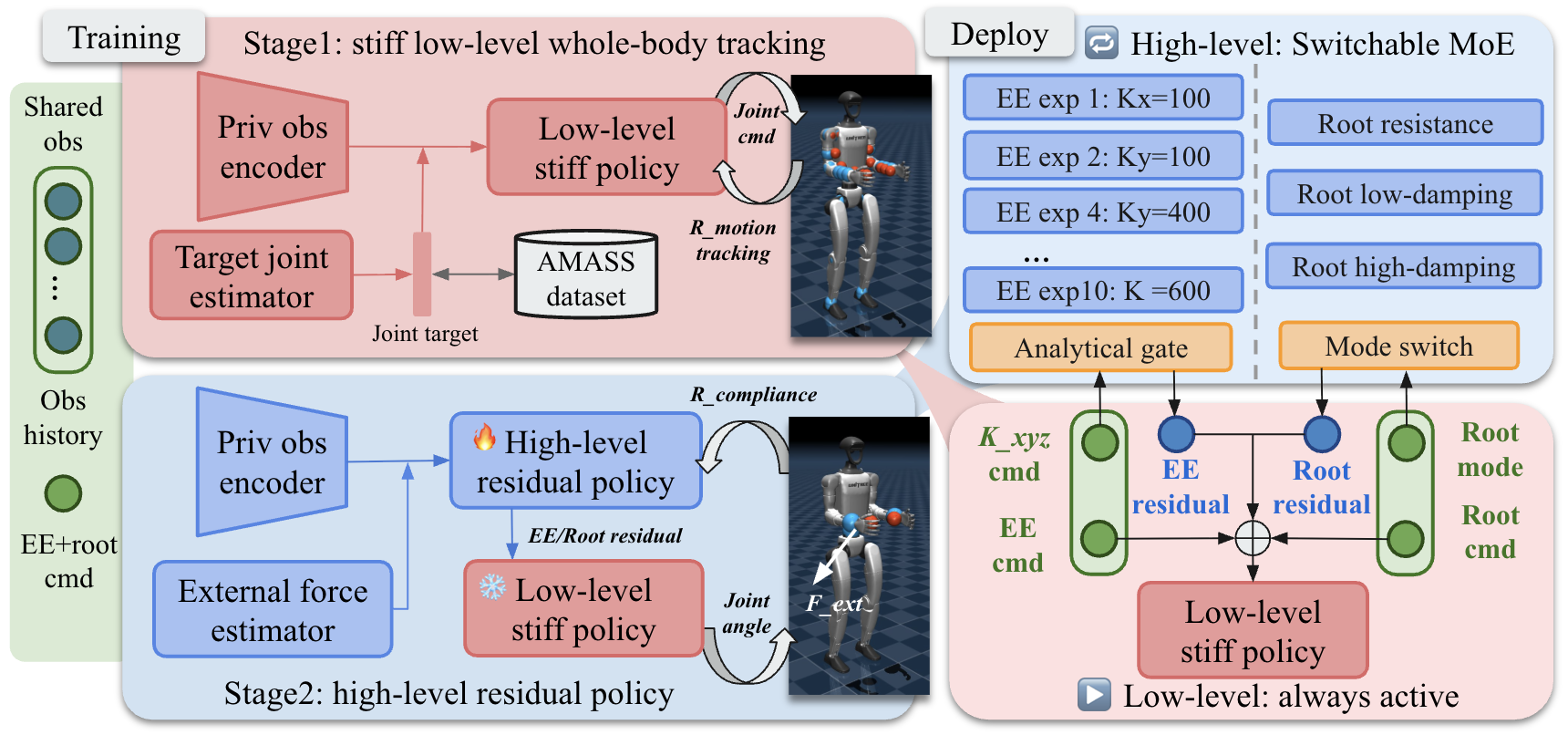}
    \caption{
System overview. Stage 1 trains a stiff low-level whole-body tracking policy with an EE--root command interface. In Stage 2, the low-level policy is held fixed while high-level residual policies learn EE or root command corrections that produce the desired response to external forces under commanded stiffness or a selected root mode. At deployment, an analytical compliance-space gate composes 10 EE compliance experts according to the commanded Cartesian stiffness, while a root compliance policy is selected through mode switching. The low-level policy remains active throughout, enabling online adjustment of EE stiffness and root mode switching.
}
    \label{fig:system_overview}
\vspace{-5mm}
\end{figure*}

We first define directional and tunable EE compliance. We define locomotion compliance in terms of root resistance and
damping modes. Then, we describe the two-stage training procedure and the
deployment-time composition and switching of the high-level compliance
policies (Fig.~\ref{fig:system_overview}).

\subsection{Directional and Tunable EE Compliance}
We define directional EE compliance for each EE $h\in\{L,R\}$ through the desired Cartesian
displacement under an external force. 
This force is 
expressed in the robot root frame as
$\mathbf{f}^{b}_{h,t}=(f^b_{h,x,t},f^b_{h,y,t},f^b_{h,z,t})$. Given the
nominal EE position command $\mathbf{x}^{\mathrm{nom},b}_{h,t}$ and commanded
Cartesian stiffness $\mathbf{K}=\operatorname{diag}(K_x,K_y,K_z)$, the target displacement is
$\Delta\mathbf{x}^{*,b}_{h,t}=(f^b_{h,x,t}/K_x,f^b_{h,y,t}/K_y,
f^b_{h,z,t}/K_z)$ and
$\mathbf{x}^{*,b}_{h,t}=\mathbf{x}^{\mathrm{nom},b}_{h,t}+
\Delta\mathbf{x}^{*,b}_{h,t}$.

\subsection{Root Resistance and Damping Modes}
We define locomotion compliance through the root frame's motion in response to external forces. We implement three root compliance modes: a resistance mode that rejects external forces, a low-damping mode that performs force following with damping of $B=60\,\mathrm{N\,s/m}$, and a high-damping mode of force following with $B=200\,\mathrm{N\,s/m}$.
For the damping mode, 
given the nominal root velocity command
$\mathbf{v}^{\mathrm{nom},w}_{xy,t}$, damping coefficient $B$, and net force
$\mathbf{f}^{w,\mathrm{net}}_{xy,t}$, the target world-frame velocity is
$\mathbf{v}^{*,w}_{xy,t}=\mathbf{v}^{\mathrm{nom},w}_{xy,t}+
\mathbf{f}^{w,\mathrm{net}}_{xy,t}/B$. These relations define the compliance targets
used to train the high-level residual policies.
Note that we use
damping rather than stiffness to define the force following behavior: the robot should keep moving while being pushed or pulled, rather than stopping at an offset position relative to a fixed equilibrium. This formulation also aligns with the low-level policy’s planar velocity command interface.

\subsection{Stage 1: Low-Level Whole-Body Tracking Policy Training}

The low-level policy is a stiff whole-body tracking controller that serves as
the backbone for high-level compliance control. It receives robot-state
observations and an EE--root target and outputs joint actions that coordinate
the whole-body to track these targets while maintaining stability.

We train the low-level policy with a DeepMimic-style dense motion-tracking
objective~\cite{deepmimic} and a three-phase teacher--student
pipeline~\cite{gentlehumanoid}. We adopt the EE--root command interface
and its corresponding tracking objectives~\cite{luo2026ceer}.
During training, external disturbance forces are applied at upper-body links while the
original motion references remain unchanged, encouraging the policy to track
stiffly under perturbations. We curate AMASS~\cite{amass} using the LIMMT data
screening criteria~\cite{guan2026limmtmotiontracking} and train a general
whole-body tracking policy conditioned on EE and root commands. Our default
three-keypoint (3-kp) interface commands the root and two EEs; the
five-keypoint (5-kp) interface commands the root, two EEs, and two feet.

\subsection{Stage 2: High-Level Residual Policy Training}


To achieve desired compliance control behaviors, the high-level compliance policies output EE or/and root command residuals, which are added to the nominal EE–root command to form the control targets provided as input to the low-level whole-body tracking policy.

\subsubsection{Residual Actions and Training Objectives}

In Stage 2, the low-level policy is frozen, and each high-level residual policy
is trained with PPO. The EE residual action consists of six Cartesian position offsets, with one $xyz$ offset for each EE. Each normalized action is passed through
$\tanh$ and scaled by an expert-specific per-axis limit before it is added to the nominal EE command. The $100\,\mathrm{N/m}$ soft-axis experts use a
$0.35\,\mathrm{m}$ limit to cover the required displacement range; the other
experts use $0.15\,\mathrm{m}$. Root compliance policies output root-command residuals; feet and yaw residuals are optional action-space extensions. Let
$\mathbf{x}^{b}_{h,t}$, $\mathbf{v}^{w}_{xy,t}$, and
$\mathbf{p}^{w}_{xy,t}$ denote the actual EE position, root planar velocity,
and root planar position, respectively, with $\mathbf{p}^{w}_{xy,0}$ denoting
the initial root position. Superscripts $b$ and $w$ indicate the robot root and
world frames, and $xy$ indicates planar components. The task errors and rewards
are
\begin{equation}
\begin{aligned}
e_t^{\mathrm{EE}} &= \tfrac{1}{2}\sum_{h\in\{L,R\}}
\|\mathbf{x}^{b}_{h,t}-\mathbf{x}^{*,b}_{h,t}\|_2,\\
e_t^{\mathrm{D}} &= \|\mathbf{v}^{w}_{xy,t}-\mathbf{v}^{*,w}_{xy,t}\|_2,\\
e_t^{\mathrm{R}} &= \|\mathbf{p}^{w}_{xy,t}-\mathbf{p}^{w}_{xy,0}\|_2,\\
r_t^{m} &= \exp(-e_t^{m}/\sigma_m),
\quad m\in\{\mathrm{EE},\mathrm{D},\mathrm{R}\}.
\end{aligned}
\end{equation}
Where $D$ and $R$ denote damping and resistance, $\sigma_m>0$ is the error
scale for objective $m$, and the factor $1/2$ averages over the two EEs. The full objective also penalizes large and temporally non-smooth command residuals.

\subsubsection{EE Compliance Experts}

An EE compliance expert receives the nominal EE command, recent commands and
actions, and robot-state history. It changes only the EE position; the nominal
EE orientation remains in the low-level command. We train 10 experts: one
isotropic $600\,\mathrm{N/m}$ expert and directional $100$, $200$, and
$400\,\mathrm{N/m}$ experts for each Cartesian axis. 
We train a separate stiffness-specific expert because our experiments showed that a single policy trained across multiple stiffness tends to produce similar force–displacement responses despite different commanded stiffness.

\subsubsection{Root Compliance Policies}

We train three root compliance policies: one resistance policy that minimizes
planar root displacement and two damping policies that track force-dependent
velocities with $B=60$ and $B=200\,\mathrm{N\,s/m}$. Each is trained as a
separate high-level residual policy.

\subsubsection{Teacher--Student Training}

In Stage 2, teacher-student training proceeds in two phases. First, a
teacher actor is trained with privileged force and simulation-state observations, while a privileged-latent estimator learns from deployable
observation histories. The teacher actor is then copied to the student
and kept fixed while this estimator continues to match the teacher latent. In
addition to the standard privileged-latent estimator, we train a
force-estimation head with direct force supervision. For the EE policies, this
head predicts the two root-frame hand forces; for the root policies, it
predicts the net force and a body-location label. At deployment, the student
uses only its observation history and the estimated latent and force features.

\subsection{Deployment-Time Policy Composition}
At deployment, an analytical compliance-space MoE composes the EE compliance
experts according to the commanded Cartesian stiffness, while a root-mode
command selects the resistance, low-damping, or high-damping policy. This
deployment scheme enables online EE stiffness adjustment, root-mode switching,
and combined EE--root compliance through the fixed low-level policy.
\subsubsection{Analytical Compliance-Space MoE}

At deployment, the analytical MoE composes the EE experts trained in Stage 2.
For a commanded stiffness $K_d$ on axis $d$, we define compliance as
$C_d=1/K_d$. Since the ideal displacement satisfies
$\Delta x_d=f_d/K_d=f_dC_d$, the outputs of the two neighboring anchor experts
are interpolated linearly in compliance space. The composed output is
$\Delta x^{\mathrm{MoE}}_{h,d}=\sum_{k\in\mathcal{N}_d}w_{d,k}
\Delta x^{(d,k)}_{h,d}$, where $\mathcal{N}_d$ contains the neighboring
stiffness anchors and the weights $w_{d,k}$ sum to one. Each Cartesian axis is
composed independently, enabling separate commands for $K_x$, $K_y$, and
$K_z$. The $100$--$600\,\mathrm{N/m}$ range is set by the apparent stiffness
of the low-level policy and the achievable residual displacement
(Sec.~\ref{sec:backbone_characterization}). Because the interpolation weights
are determined directly by $C_d=1/K_d$, the gate has no learned parameters.

\subsubsection{Root Policy Selection}

At deployment, a root-mode command selects one of the three policies trained
in Stage 2. Unlike the EE experts, the root policies are not interpolated
because the root response depends on foot--ground contacts and support
conditions.

\subsubsection{Combined EE--Root Control}

The analytical MoE adds an EE command residual and the selected root policy
adds a root command residual to the nominal EE--root command. Because the two
residuals modify different command components, EE and root compliance can run
simultaneously without a priority rule or policy reset.

%% file: experiments.tex
\section{Experiments}
\label{sec:experiments}

Our experiments address three main questions:
(i) how accurately the proposed framework realizes commanded directional EE
stiffness and root compliance modes; (ii) whether the two-level hierarchy and
analytical compliance-space MoE improves accurate directional control over the full stiffness range; and (iii) whether and why the combination of directional and tunable EE compliance with root compliance improve real-world loco-manipulation tasks.

We first describe the perturbation protocol, low-level policy, and force and
stiffness ranges. We then evaluate the EE hierarchy, full-range compliance,
and workspace consistency, followed by root modes and action spaces.
Finally, we demonstrate combined EE--root compliance in real-world
loco-manipulation.

\subsection{Experimental Setup and Evaluation Protocol}
\label{sec:experimental_setup}

All policies are trained in Isaac Sim with 16,384 parallel environments on
eight L40S GPUs. The low-level (LL) policy uses 7 billion environment steps and
takes approximately 6 hours. Each high-level (HL) policy uses 5 billion environment steps
without fine-tuning and takes approximately 2 hours. The stiffness-conditioned
HL baseline is trained for 10 billion environment steps to ensure convergence. Real-world
experiments use a Unitree G1, with policy inference running on an RTX 4080
laptop connected to the robot via Ethernet. No wrist force/torque sensors are
used.

We evaluate EE compliance in simulation. 
In each trial, each EE is commanded to a fixed initial pose.
External forces are independently applied to the corresponding left and right hands
along the positive and negative $x$-, $y$-, and $z$-directions, with magnitudes
of $\{5,10,15,20,30\}~\mathrm{N}$. This results in $2$ hands $\times$ $3$
axes $\times$ $2$ directions $\times$ $5$ force magnitudes $=60$ trials for
each evaluated stiffness configuration. Each force is ramped from zero over
$0.5~\mathrm{s}$ and then held for $2.0~\mathrm{s}$, giving $2.5~\mathrm{s}$
of force exposure. Measurements are averaged over the final $0.5~\mathrm{s}$
of the hold period.

\textbf{EE compliance-tracking error} is the Euclidean distance between the
actual EE position and the compliant target. \textbf{Nominal EE tracking
error} is the Euclidean distance between the actual EE position and the
nominal EE command. \textbf{Apparent stiffness} is estimated from the
ratio of the applied force to the resulting EE displacement along each
Cartesian axis. Additional
range metrics are introduced in Sec.~\ref{sec:range_compliance}.

All learned policies are evaluated as deployable students without privileged
inputs. Unless otherwise specified, all hierarchical methods use the same stiff 3-kp
low-level policy.

\subsection{Low-Level Policy Characterization}
\label{sec:backbone_characterization}

Before evaluating the high-level controller, we characterize our stiff 3-kp low-level backbone against two representative alternatives: a compliant policy CEER~\cite{luo2026ceer} and a general motion-tracking policy SONIC~\cite{luo2026sonic}. We also vary the maximum training force of the 3-kp policy from $30$ to $60~\mathrm{N}$. The results identify our 30-N 3-kp policy as the most suitable backbone, because it has low nominal tracking error with high apparent stiffness.

\begin{table}[t]
\vspace{2mm}
\centering
\caption{Low-level tracking-policy and training-force characterization.
Apparent stiffness values describe natural behavior and are not errors
relative to a target.}
\label{tab:backbone_characterization}
\small
\setlength{\tabcolsep}{3.0pt}
\resizebox{\columnwidth}{!}{%
\small
\begin{tabular}{lcccc}
\toprule
\textbf{Policy} &
\shortstack{\textbf{Nom. Err.} (m) $\downarrow$} &
\shortstack{\textbf{Median} (N/m)} &
\shortstack{\textbf{P95} (N/m)} \\
\midrule
3-kp stiff (30 N) & $0.0456 \pm 0.0140$ & 621.4 & 1793.0 \\
CEER\cite{luo2026ceer}             & $0.1385 \pm 0.1657$ & 103.3 & 285.0 \\
SONIC\cite{luo2026sonic}            & $0.0955 \pm 0.0482$ & 192.3 & 342.4 \\
\midrule
3-kp stiff (40 N) & $0.0524 \pm 0.0148$ & 544.7 & 961.8 \\
3-kp stiff (50 N) & $0.0528 \pm 0.0134$ & 672.1 & 3426.0 \\
3-kp stiff (60 N) & $0.0830 \pm 0.0269$ & 406.5 & 1088.8 \\
\bottomrule
\end{tabular}
}%
\end{table}

As shown in Table~\ref{tab:backbone_characterization}, the stiff 3-kp policy
provides the lowest nominal error and a median apparent stiffness near
$600~\mathrm{N/m}$, making it a suitable low-level policy for high-level
compliance training. Increasing the
maximum training force from $30$ to $60~\mathrm{N}$ does not monotonically
increase apparent stiffness and degrades nominal tracking, most clearly at
$60~\mathrm{N}$. We therefore use the 30 N policy in the main experiments.

\begin{figure}[t]
    \centering
    \includegraphics[width=\columnwidth]{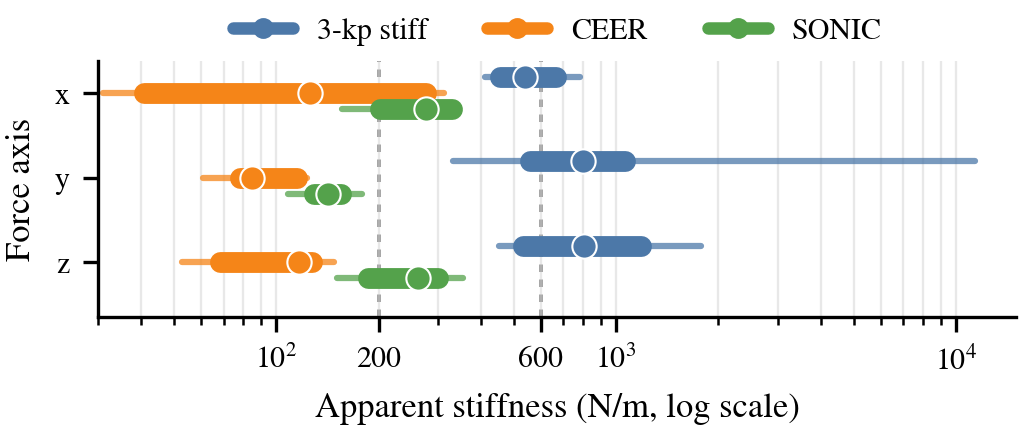}
    \caption{Natural apparent stiffness of representative low-level policies.
    Positive and negative directions are combined for each axis. Thin lines show
    P5--P95, thick lines show the interquartile range, and dots show the median.
    Dotted lines indicate $200$ and $600~\mathrm{N/m}$.}
    \label{fig:low_level_stiffness}
\end{figure}

Figure~\ref{fig:low_level_stiffness} compares the natural Cartesian response of
the three representative low-level policies, with measurements grouped by
perturbation axis for clarity. Their apparent-stiffness distributions differ
substantially across policies. CEER is naturally more compliant, while SONIC exhibits an intermediate
response. The broad, uncommanded stiffness distributions motivate explicit
high-level compliance control.

We select $600~\mathrm{N/m}$ as the upper command bound because it is close to
the overall median apparent stiffness of the low-level policy. The lower bound
is constrained by reachable EE displacement: the curated AMASS motion
corpus~\cite{amass} has P5--P95 EE position widths above $0.4~\mathrm{m}$,
while $K=100~\mathrm{N/m}$ under the
maximum evaluated force of $30~\mathrm{N}$ requires a $0.30~\mathrm{m}$
displacement. We therefore set the soft-axis residual limit to
$0.35~\mathrm{m}$, providing a $0.05~\mathrm{m}$ margin while remaining within
the observed EE range. We use the $100$--$600~\mathrm{N/m}$ command range,
treating $100~\mathrm{N/m}$ as its workspace-limited lower bound.

\subsection{End-Effector Compliance Evaluation}
\label{sec:ee_compliance_evaluation}

We evaluate hierarchy design, full-range directional compliance, and workspace consistency. The results show that the hierarchy reduces compliance-tracking error, the analytical MoE yields the lowest full-range errors, and errors remain similar across the tested EE poses.

\subsubsection{Fixed-Target Hierarchy Ablation}
\label{sec:policy_external_force}

We first evaluate whether separating compliance control from whole-body
motion tracking improves accuracy at a single target stiffness. All methods in
this experiment are trained or commanded to realize an isotropic stiffness of
$200~\mathrm{N/m}$.

\textbf{Analytical MoE} is the full proposed method.
\textbf{Fixed-stiffness HL} replaces the expert composition with a single
high-level residual policy trained only for the $200~\mathrm{N/m}$ target.
\textbf{Fixed-stiffness E2E} removes the hierarchy and directly trains a
single whole-body policy for the same target.

\begin{table}[t]
\vspace{2mm}
\centering
\caption{
Fixed-stiffness compliance evaluation at $200~\mathrm{N/m}$.
}
\label{tab:ee_compliance_eval}
\small
\setlength{\tabcolsep}{3.5pt}
\resizebox{\columnwidth}{!}{%
\begin{tabular}{lccc}
\toprule
\textbf{Policy} &
\shortstack{\textbf{EE Comp. Err.} \\ (m) $\downarrow$} &
\shortstack{\textbf{Nom. EE Err.} \\ (m) $\downarrow$} &
\shortstack{\textbf{App. Stiffness} \\ (N/m)} \\
\midrule

Analytical MoE
& $\mathbf{0.023 \pm 0.016}$
& $0.049 \pm 0.045$
& $219.6 \pm 49.2$ \\

Fixed-stiffness HL
& $0.028 \pm 0.014$
& $0.050 \pm 0.040$
& $241.5 \pm 55.7$ \\

Fixed-stiffness E2E
& $0.082 \pm 0.034$
& $0.100 \pm 0.046$
& $316.6 \pm 512.0$ \\

\bottomrule
\end{tabular}
}%
\end{table}

Table~\ref{tab:ee_compliance_eval} shows that the analytical MoE and
fixed-stiffness HL policy achieve EE compliance-tracking errors of
$0.023 \pm 0.016~\mathrm{m}$ and
$0.028 \pm 0.014~\mathrm{m}$, respectively, compared with
$0.082 \pm 0.034~\mathrm{m}$ for fixed-stiffness E2E. The two hierarchical
methods also achieve apparent stiffness values closer to $200~\mathrm{N/m}$.
The hierarchy therefore improves
fixed-stiffness EE compliance, while analytical compliance-space composition preserves the
accuracy of a dedicated fixed policy.

\subsubsection{Full-Range Directional EE Compliance}
\label{sec:range_compliance}

We evaluate 16 stiffness configurations spanning the complete command range:
three settings with one soft axis at $100~\mathrm{N/m}$ and the other two at
$600~\mathrm{N/m}$; three settings with one stiff axis at $600~\mathrm{N/m}$
and the other two at $100~\mathrm{N/m}$;
four isotropic settings at $100$, $200$, $400$, and $600~\mathrm{N/m}$; and
six permutations of $150$, $300$, and $500~\mathrm{N/m}$ across the Cartesian
axes.
These settings test cross-axis expert composition, with the six non-anchor
permutations additionally testing interpolation on all three axes.
For each configuration, we follow the perturbation protocol in
Sec.~\ref{sec:experimental_setup}, resulting in 60 trials per configuration
and 960 trials for each method in total.

\paragraph{Baselines}
\textbf{Stiffness-conditioned HL} replaces the analytical MoE with one
stiffness-conditioned high-level residual policy over the same stiff low-level
policy. \textbf{Stiffness-conditioned E2E} trains a single
stiffness-conditioned whole-body
policy without the hierarchy. Both learned baselines are trained until
convergence. The \textbf{oracle-force LL} baseline uses the ground-truth
force to compute an EE command offset,
$\Delta\mathbf{x}^{*}=\mathbf{K}^{-1}\mathbf{F}_{\mathrm{ext}}$, which is added
to the nominal EE command. It tests whether analytical force-to-offset
conversion alone is sufficient when force-estimation error is removed.

\paragraph{Metrics}
For trial $i$, apparent stiffness is
$\hat K_i=\lVert\mathbf F_i\rVert/
|\mathbf d_i^\top\overline{\Delta\mathbf p}_i|$, where $\mathbf d_i$ is the
unit applied-force direction and $\overline{\Delta\mathbf p}_i$ is the EE
displacement from the unperturbed baseline. Stiffness MAPE is
$N^{-1}\sum_i|\hat K_i-K_i^{\mathrm{cmd}}|/K_i^{\mathrm{cmd}}$, where
$N=960$ is the number of trials per method and $K_i^{\mathrm{cmd}}$ is the
commanded stiffness along the applied-force direction.
To capture directional coupling, we fit
$\Delta\mathbf x=\mathbf b+\hat{\mathbf C}\mathbf F$ by least squares and
report the normalized compliance-matrix error
$\lVert\hat{\mathbf C}-\mathbf C^*\rVert_F/\lVert\mathbf C^*\rVert_F$, where
$\mathbf C^*=\operatorname{diag}(1/K_x,1/K_y,1/K_z)$; this error is averaged
over both hands and all stiffness configurations. Finally, a global
ordinary-least-squares fit with an intercept,
$\hat K_{\mathrm{measured}}=\alpha K_{\mathrm{commanded}}+\beta$, gives the
trend slope $\alpha$ and $R^2$, both ideally $1$.

\begin{figure}[t]
\vspace{2mm}
    \centering
    \includegraphics[width=\columnwidth]{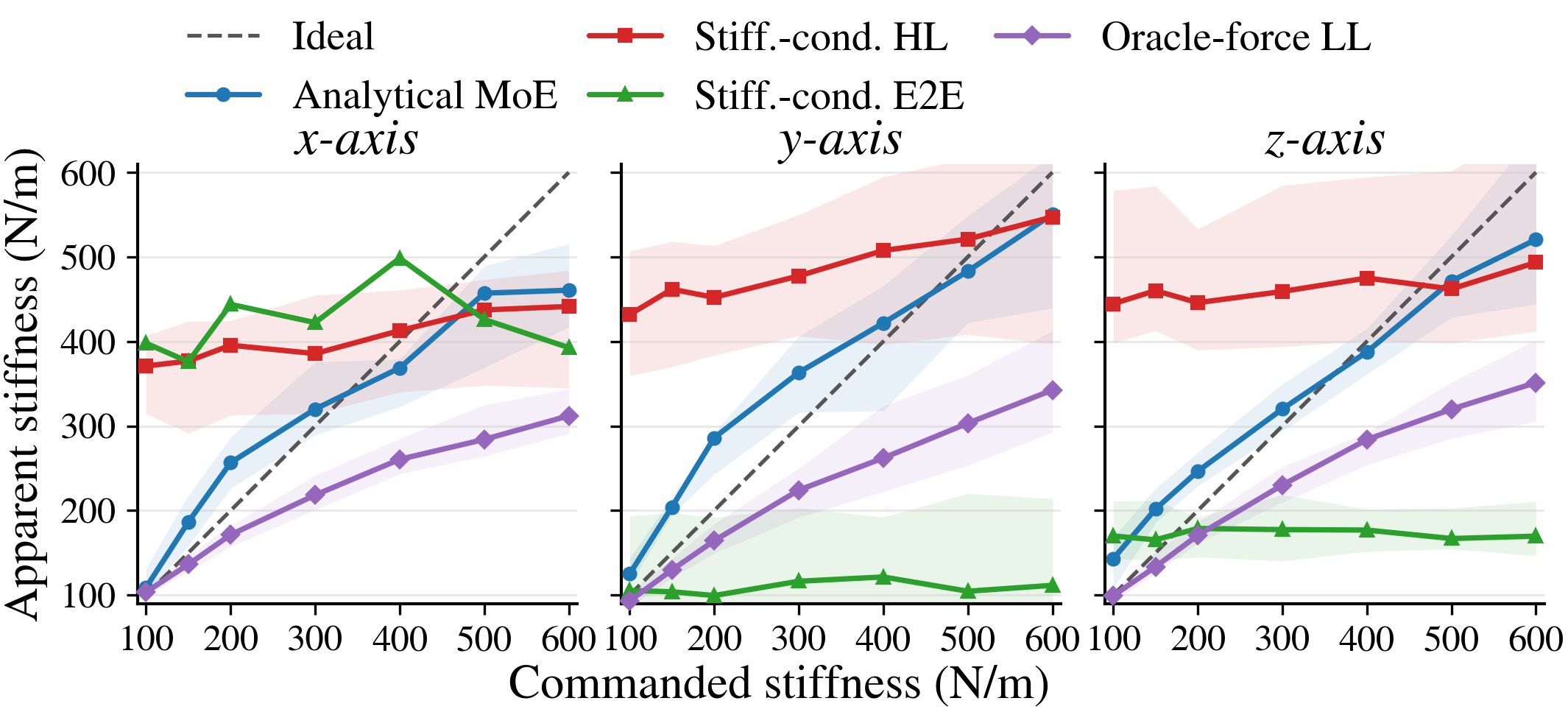}
    \caption{Commanded versus apparent stiffness over the 16-configuration
    100--600\,N/m evaluation, shown separately for the $x$, $y$, and $z$ axes.
    Curves show the median response, shaded regions show the interquartile
    range, and the dashed line denotes ideal tracking. For readability, the
    unusually broad stiffness-conditioned E2E IQR band is omitted only in the $x$ panel.}
    \label{fig:range_command_response}
\end{figure}

\paragraph{Results}
Table~\ref{tab:range_compliance_eval} shows that the analytical MoE achieves the
lowest compliance-matrix error ($0.278$), EE compliance-tracking error
($0.0349~\mathrm{m}$), and stiffness MAPE ($0.249$), with the slope closest to
$1$. As shown in Fig.~\ref{fig:range_command_response}, the
stiffness-conditioned HL policy maps the full command range to much narrower,
axis-dependent response ranges. Its low global slope ($0.136$) quantifies this
range compression, while its low $R^2$ ($0.040$) indicates poor
command--response consistency. The stiffness-conditioned E2E policy further
degrades accuracy and has nearly zero $R^2$.

This range compression is consistent with multi-task interference: a
single stiffness-conditioned policy can optimize a compromise response rather
than preserve distinct axis-specific behaviors across the command range.
End-to-end training further couples compliance learning with whole-body motion
tracking. The analytical MoE separates these objectives and composes
specialized EE compliance behaviors, reducing both sources of interference.

The oracle-force LL baseline produces a consistent trend ($R^2=0.792$), but its
slope of $0.485$ and EE compliance-tracking error of $0.0459~\mathrm{m}$ remain
worse than those of the analytical MoE.
Directly applying $\mathbf{K}^{-1}\mathbf{F}_{\mathrm{ext}}$ assumes that the
low-level policy is an ideal Cartesian position controller. Its finite,
unregulated apparent stiffness and coupled whole-body dynamics violate this
assumption, so the requested displacement is not realized exactly. Notably, our method
outperforms the oracle-force baseline despite using estimated rather than
ground-truth forces, providing evidence that the learned EE compliance experts
\emph{compensate} for the low-level closed-loop dynamics, not merely infer the
external force.

\begin{table}[t]
\vspace{2mm}
\centering
\caption{
Full-range directional EE compliance over 16 stiffness configurations
(ideal trend slope and $R^2$: $1$).
}
\label{tab:range_compliance_eval}
\small
\setlength{\tabcolsep}{3.2pt}
\resizebox{\columnwidth}{!}{%
\begin{tabular}{lccccc}
\toprule
& \multicolumn{3}{c}{\textbf{Accuracy}} &
\multicolumn{2}{c}{\textbf{Trend}} \\
\cmidrule(lr){2-4}
\cmidrule(lr){5-6}

\textbf{Method} &
\shortstack{\textbf{Compliance Matrix}\\\textbf{Err.} $\downarrow$} &
\shortstack{\textbf{EE Comp. Err.}\\(m) $\downarrow$} &
\shortstack{\textbf{Stiffness}\\\textbf{MAPE} $\downarrow$} &
\shortstack{\textbf{Slope}\\$\rightarrow 1$} &
\shortstack{\textbf{$R^2$}\\$\uparrow$} \\
\midrule

Analytical MoE
& $\mathbf{0.278}$
& $\mathbf{0.0349}$
& $\mathbf{0.249}$
& $\mathbf{0.764}$
& 0.779 \\

Stiff.-cond. HL
& 0.677
& 0.0566
& 1.325
& 0.136
& 0.040 \\

Stiff.-cond. E2E
& 1.064
& 0.0884
& 5.208
& 3.430
& 0.003 \\

Oracle-force LL
& 0.304
& 0.0459
& 0.256
& 0.485
& $\mathbf{0.792}$ \\

\bottomrule
\end{tabular}
}%
\end{table}

\subsubsection{Workspace Consistency}
\label{sec:workspace_consistency}

We evaluate workspace consistency at seven paired EE poses: the center pose
and $0.05~\mathrm{m}$ offsets along each positive and negative Cartesian axis.
At each pose, we apply the full 16-configuration protocol, totaling 6,720
trials. For each hand, pose, and Cartesian component, we average the absolute
difference between the actual and target EE displacements over 480 trials.
Across the tested poses, the mean $x$-, $y$-, and $z$-component errors are $24.8$,
$22.1$, and $25.8~\mathrm{mm}$, respectively. The largest pose-wise mean
component error is $43.7~\mathrm{mm}$ at the raised-hand $z+$ pose, which may
reflect reduced joint-limit margin or kinematic redundancy when the arms are
elevated. The two hands show similar qualitative trends but retain quantitative
asymmetries: the left $x$ response is generally less accurate, whereas the right
$z$ response is generally less accurate.

\subsection{Root Compliance Evaluation}
\label{sec:locomotion_compliance}

We next evaluate whether the root compliance policies produce distinct
resistance and damping responses. Forces of
$\{5,10,20,30\}~\mathrm{N}$ are applied along four horizontal directions at
the torso, shoulders, wrists, and hands, giving 112 cases per policy. We study
three modes. \textbf{Resistance} minimizes root displacement and uses zero target
velocity. The two damping modes target an along-force velocity change
$\Delta\mathbf{v}^{*}=\mathbf{F}_{xy}/B$ with $B=60$ or
$200~\mathrm{N\,s/m}$. Planar root drift measures position retention, while
$\lVert\Delta\mathbf{v}-\Delta\mathbf{v}^{*}\rVert$ measures the root
velocity-response error.

\begin{table}[t]
\centering
\caption{Root-compliance evaluation using the 3-kp root-residual action space
without yaw control.}
\label{tab:root_compliance}
\small
\setlength{\tabcolsep}{4.0pt}
\resizebox{\columnwidth}{!}{%
\begin{tabular}{lccc}
\toprule
\textbf{Mode} &
\shortstack{\textbf{Root drift}\\(m) $\downarrow$} &
\shortstack{\textbf{Velocity error}\\(m/s) $\downarrow$} &
\shortstack{\textbf{Apparent $B$}\\median / target} \\
\midrule
3-kp LL only & 0.3627 & 0.0955 & -- \\
Resistance & 0.3103 & 0.0987 & -- \\
$B=60$ & 1.7806 & 0.1143 & 57.0 / 60 \\
$B=200$ & 0.7163 & 0.1031 & 201.4 / 200 \\
\bottomrule
\end{tabular}
}%
\end{table}

Table~\ref{tab:root_compliance} shows that the three root compliance modes
yield distinct responses. The resistance mode moves least, $B=60$ follows the
pull most, and $B=200$ gives an intermediate response. The median apparent
damping closely matches both commanded values.

\begin{table}[t]
\vspace{2mm}
\centering
\caption{Root-policy action-space ablation. Resistance is evaluated by planar
root drift (m), and damping by velocity-response error (m/s). All variants include root-command residuals; 5-kp adds foot-position residuals, and +yaw adds a root-yaw residual.}
\label{tab:root_action_ablation}
\small
\setlength{\tabcolsep}{3.2pt}
\resizebox{0.80\columnwidth}{!}{%
\begin{tabular}{lcccc}
\toprule
\textbf{Mode} & \textbf{3-kp} & \textbf{3-kp+yaw} &
\textbf{5-kp} & \textbf{5-kp+yaw} \\
\midrule
Resistance  & 0.3103 & 0.2032 & 0.4964 & $\mathbf{0.1574}$ \\
$B=60$  & 0.1143 & 0.1217 & 0.1094 & $\mathbf{0.1073}$ \\
$B=200$ & 0.1031 & $\mathbf{0.0678}$ & 0.0904 & 0.0700 \\
\bottomrule
\end{tabular}
}%
\end{table}
\vspace{-1mm}

We further test whether additional locomotion-relevant control authority
improves the three root modes. As shown in
Table~\ref{tab:root_action_ablation}, foot and yaw residuals substantially
improve the resistance mode, where active turning and stepping reject
disturbances. Their improvements in the damping modes are smaller and
configuration-dependent. Thus, additional control authority helps
stabilization but does not fundamentally resolve contact- and gait-dependent
damping control.

\subsection{Real-World Evaluation of EE and Root Compliance}
\label{sec:real_world_tasks}

We directly transfer all policies from simulation to real and evaluate tunable Cartesian stiffness, directional
EE compliance, and combined EE--root compliance in real-world tasks,
including online adjustment of EE stiffness. Because manual force and displacement measurements yield only approximate stiffness estimates, we focus primarily on task-level performance.
Fig.~\ref{fig:teaser} provides an overview, while Fig.~\ref{fig:exp_demo}
summarizes the trials. Complete experiments appear in the supplementary video
and on the project website.

\subsubsection{Tunable EE Compliance and Online Stiffness Adjustment}
We estimate the real-world end-effector stiffness by pulling the EE with a force gauge. The ruler reading changes from $1$ to $3\,\mathrm{in}$,
giving a net displacement of $2\,\mathrm{in}$. Under $K_y=100$, $200$, and
$600\,\mathrm{N/m}$, the measured forces are $10.9$, $17.2$, and
$28.0\,\mathrm{N}$, respectively, corresponding to apparent
stiffness values of $214.6$, $338.6$, and $551\,\mathrm{N/m}$
(Fig.~\ref{fig:exp_demo}(m--o)). The remaining discrepancy is reasonable given the uncertainty in the manual force and displacement measurements. In yoga-ball grasping, soft and stiff $y$-axis settings produce visibly
different ball indentations (Fig.~\ref{fig:exp_demo}(p,q)). In payload
dragging, increasing $K_z$ reduces the vertical EE displacement under the same
$3\,\mathrm{kg}$ payload (Fig.~\ref{fig:teaser}(b)). The EE stiffness is
adjusted online without restarting the low-level policy.

\subsubsection{Directional EE Compliance}
\begin{figure}[!t]
    \centering
\includegraphics[width=\columnwidth]{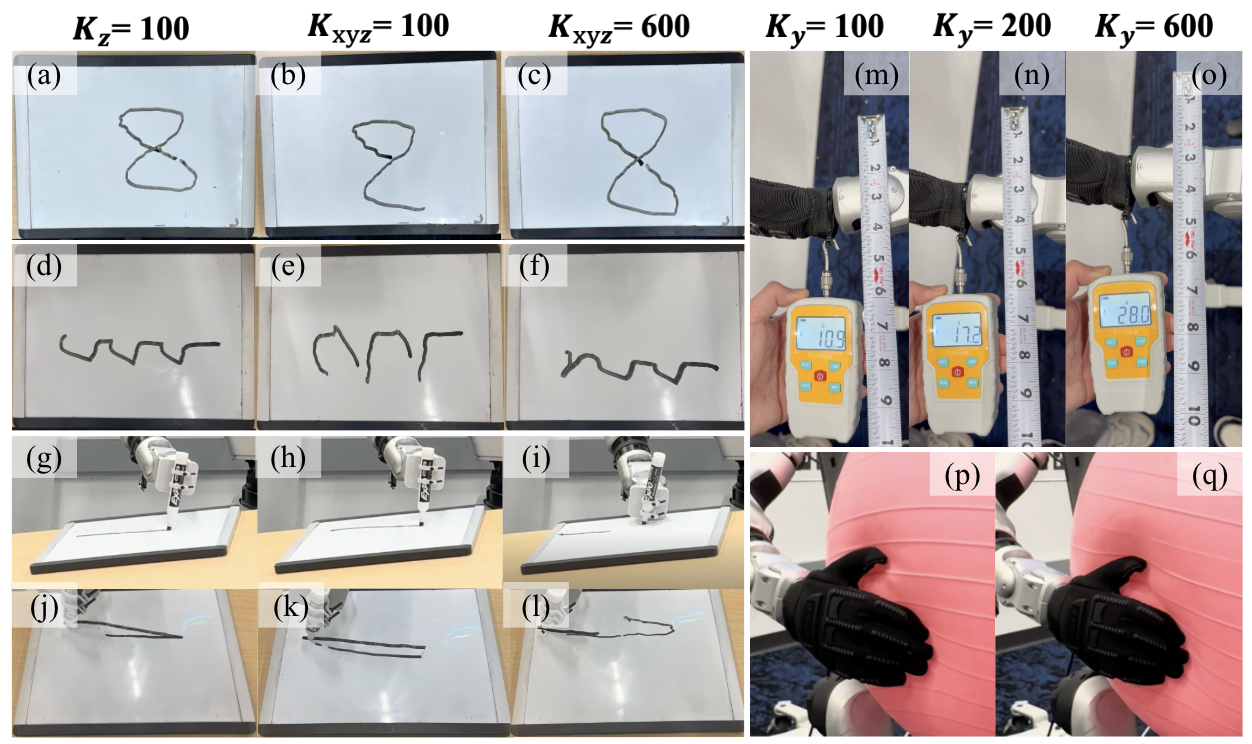}
    \caption{Directional EE compliance tasks: writing ``8'' (a--c), straight-line writing under disturbance (d--f), and writing on a tilted surface (g--l). Tunable EE-stiffness demonstrations: force-gauge pulling (m--o) and yoga-ball grasping (p--q).}
    \label{fig:exp_demo}
\end{figure}

Writing requires low surface-normal stiffness $K_z$ and high tangential
stiffness $(K_x,K_y)$. When writing ``8'' (Fig.~\ref{fig:exp_demo}(a--c)), the
isotropically soft policy often loses contact, whereas the directional
$K_z=100\,\mathrm{N/m}$ and isotropically stiff settings complete the trajectory.
The isotropically stiff setting, however, breaks the pen holder in another trial
(Fig.~\ref{fig:teaser}(c)). We then draw a straight line while applying a
disturbance perpendicular to the desired line (Fig.~\ref{fig:exp_demo}(d--f)).
Increasing stiffness along the disturbance direction visibly reduces the
resulting protrusion in the drawn trajectory. On a surface tilted by
$10^\circ$, panels (g) and (h) compare whether the pen tilts under different
directional settings, while the line in panel (j) is less straight than that
in panel (k). Isotropically high stiffness again breaks the holder. Together, these results demonstrate the need for directional EE compliance: low surface-normal stiffness preserves safe contact, while high tangential
stiffness improves trajectory accuracy.

\subsubsection{Combined Directional EE and Root Compliance}
We verified the usefulness of combined directional EE and root compliance via collaborative box carrying, a human-robot co-manipulation task.
The EE is compliant along the contact $y$-axis
and stiff along $x$ and $z$, while the root uses the $B=60$ damping mode to follow the human interaction forces. With a $3\,\mathrm{kg}$ payload, this combination completes the task, whereas isotropically low EE stiffness cannot maintain support and the box slips
(Fig.~\ref{fig:teaser}(a)). This task requires directional EE compliance to accommodate interaction while supporting the payload and root damping to follow the human, highlighting the need for combined EE--root compliance.

